# A novel information gain-based approach for classification anddimensionality reduction of hyperspectral images

Asma Elmaizi*[a], Hasna Nhaila [a], Elkebir Sarhrouni [a], Ahmed Hammouch [a], Chafik Nacir [a]

[a]Laboratory LRGE, ENSET, Mohammed V University Rabat, Morocco

## Abstract

Recently, the hyperspectral sensors have improved our ability to monitor the earth surface with high spectral resolution. However, the high dimensionality of spectral data brings challenges for the image processing. Consequently, the dimensionality reduction is a necessary step in order to reduce the computational complexity and increase the classification accuracy. In this paper, we propose a new filter approach based on information gain for dimensionality reduction and classification of hyperspectral images. A special strategy based on hyperspectral bands selection is adopted to pick the most informative bands and discard the irrelevant and noisy ones. The algorithm evaluates the relevancy of the bands based on the information gain function with the support vector machine classifier. The proposed method is compared using two benchmark hyperspectral datasets (Indiana, Pavia) with three competing methods. The comparison results showed that the information gain filter approach outperforms the other methods on the tested datasets and could significantly reduce the computation cost while improving the classification accuracy.

## 1. Introduction

Recently, hyperspectral image classification has been a very active field of research in many areas [1][2]. Thistechnology consists of collecting images with high resolution simultaneously in hundreds of narrow and contiguousspectral frequencies. The hyperspectral sensors wavelengths ranging can vary from the visible to the infrared spectrum [3].

The hyperspectral data provides rich and detailed observation and leads to a better materials discrimination and classification [4], which cannot be identified by multispectral images. The hyperspectral technology has been used for the first time by the National Aeronautics and Space Administration (NASA). The NASA build the AVIRIS spectrometer that monitor the earth surface and record hyperspectral image using more than 220 hyperspectral bands. Although the HSI were originally developed for remote sensing, mining and geology, it has now spread into different applications as mineralogy, food processing, agriculture and biomedical imaging. Despite all these advantages, the large number of bands and the predictive instances influence the classification accuracy and may cause the Hughes phenomenon [5] or Bellman curse of dimensionality [6].

---
* Corresponding author. Tel.: +212673256043
E-mail address: asma.elmaizi@gmail.com

The hyperspectral sensors wavelengths ranging can vary from the visible to the infrared spectrum [3]. The hyperspectral data provides rich and detailed observation and leads to a better materials discrimination and classification [4], which cannot be identified by multispectral images. The hyperspectral technology has been used for the first time by the National Aeronautics and Space Administration (NASA). The NASA build the AVIRIS spectrometer that monitor the earth surface and record hyperspectral image using more than 220 hyperspectral bands. Although the HSI were originally developed for remote sensing, mining, and geology, it has now spread into different applications as mineralogy, food processing, agriculture and biomedical imaging. Despite all these advantages, the large number of bands and the predictive instances influence the classification accuracy and may cause the Hughes phenomenon [5] or Bellman curse of dimensionality [6].

In fact, we should take into consideration that some spectral bands might be noisy and irrelevant for classification. The contiguous bands are highly correlated with each other and may contain redundant info which may cost computational burden for HSI. Those challenges may degrade the classification accuracy and increase the computational time that depends on the bands number and the pixels used for the training.

To meet the previous challenges, the dimensionality reduction is an essential step toward improving the hyperspectral images classification. The feature selection and extraction are known to be important techniques in hyperspectral images classification. Feature extraction aims at reducing the dimension of the collected data by projecting the hyperspectral image into another feature space and applying a linear transformation. The extraction techniques include unsupervised approaches such as principal component analysis (PCA) [7] and independent component analysis (ICA)[8], as well as supervised approaches such as linear discriminant analysis [9]. These methods involve transforming the data and some crucial and critical information may be compromised and distorted.

Besides feature extraction, feature selection techniques aim to find the best subset of bands from the original. In hyperspectral image analysis, band selection is preferred over band extraction for dimensionality reduction. They have the advantage of preserving the relevant original information of the data.

Band selection techniques are categorized into two groups [10]: the wrapper approaches that are classifier-dependent and the filters which are classifier-independent.

- Wrapper methods rank the bands based on the classification accuracy given by the used classifier. They have the disadvantage of being slower and computationally expensive especially with the hyperspectral data.

- Filter methods evaluate the quality of the selected bands independently of the classifier. They define a goal function that evaluate the picked bands and measure their relevancy for the classification. The filter methods are generally preferred to wrappers for being much faster, simple and efficient for HIS classification.

The topic of band selection using filter approaches has been reviewed in detail in a number of recent review articles. In the supervised learning, the filter methods rank bands according to their relevance to the class label. Many techniques have been proposed to compute the relevance score including distance, information, correlation and consistency measures. Different type of selection methods based on information-theory [11] have been proposed and will be the main topic of the work presented in this paper.

Several studies consider relevance of bands based on information theory that was proposed by (Cover & Thomas, 2006) [11]. For example, Mutual information (MI) is considered as one of the simplest approaches used for band selection. This mutual information maximization was presented by Lewis [12]. The MIM returns a ranking of features on the basis of their individual mutual information with the class MI(fs,C). Mutual Information Feature Selection (MIFS) introduced by Battiti [13] suggest that a good set of features should be highly correlated with the Class label and not correlated with each other. Battiti introduce a new criterion eliminating the redundancy between features. Peng and al [14] also proposed the minimum redundancy maximum relevance (MRMR) algorithm as an improvement of MIFS where the redundancy term is divided over the cardinality of the subset. He joins the relevancy and the redundancy in the same equation.

Sarhrouni [15][16] proposed an effective mutual information-based filter approach (MIBF). He builds an approximation of Class (ground truth) on each step of the algorithm based on the feature (band) that has the largest MI with the ground truth. Consequently, the selected bands will generate the closest estimation of the ground truth (Class).

In this paper, we propose a novel filter for bands selection based on information gain for hyperspectral dimensionalityreduction and classification. The contribution of this approach is modeling a function that takes into account two terms: relevance and interaction term based on the information gain. The information gain is more effective than theclassical mutual information and will select the relevant bands which will increase the classification accuracy. To evaluate the performance of the proposed algorithm, we present a comparison with various state-of-the-art band selection methods on two benchmark hyperspectral datasets Pavia and Indiana. The Support Vector Machine (SVM)classifier [17] [18] is used to assess the classification accuracy. Rest of the paper is organized as follows: part II describes the proposed approach. In part III, we present the datasets and the setting parameters. It also includes the experiments conducted on the Indiana and Pavia datasets and the comparison results with the other methods. Part IVconcludes the paper with major perspectives.

## 2. The proposed information gain-based approach

### 2.1. Limitations of the current band selection functions

While reviewing the methods discussed on the previous part, we can remark that most of the approaches are generally based on two criteria: relevance and redundancy. For example, the MIM algorithm is based on a ranking goal function. During each iteration, the MIM algorithm is selecting the feature that have the maximum MI with the class (groundtruth). The MIM method is simply and rapidly executed, but its main weakness is the redundancies between bands that is neglected during the selection process. The redundancy issue has been partly solved by Sarhrouni[15] and Guo[19]. Sarhrouni introduced an appropriate threshold in order to control the redundancy between the bands. A wrong choice of the redundancy criteria may lead to some limitations on the selection process.
Batiti [13] also proposed the Mutual Information Feature Selection (MIFS) filter to overcome the redundancy limitation. He suggested that a good set of bands should be highly correlated with the ground truth and not correlated with eachother. To improve the redundancy control, Peng and al [14] proposed the minimum redundancy maximum relevance(MRMR) method where the subset is divided over the cardinality.
Although, the MIFS [13] and the MRMR [14] approaches attempt to simultaneously maximize the relevancy and minimize the redundancy terms but they ignore the information gain between the picked bands . These methods don'tevaluate the class information added by each band to the already selected ones during the selection process. They assume that the irrelevant bands with a poor mutual information factor can't add any extra info for classification whencombined with the other hyperspectral bands.

### 2.2. The proposed band selection algorithm

The aim behind hyperspectral image classification is to classify each pixel of a spectral image into a specific class.Therefore, we will consider developing an approach for dimensionality reduction of spectral bands in order to getthe best classification. We denote the class as ground truth (GT), the features or bands to reduce as (B).
The basic idea of the proposed filter approach is to find a compromise between bands interaction and bandsredundancy.
The algorithm selects the discriminative bands using an evaluation function based on the mean of Interaction Gain.We build the subset adding the band that maximize the compromise between:

a) The mutual information between the selected band and the ground truth I(GT,B) in order to select the relevant and eliminate the noisy bands. The relevance is calculated by measuring the mutual information of each band with the ground truth as presented in equation 1.

$$relevance\ (B) = MI\ (B, GT) \qquad (1)$$

The mutual information measure the shared information between 2 features A and B. Theoretically, the information theory [11] defines the MI as the difference between two entropies, equation 2:

$$MI(A,B) = H(A) - H(A/B) = \sum_{A,B} P(A,B) \log_2 \frac{P(A,B)}{P(A)P(B)} dAdB \qquad (2)$$

$$H(A) = \sum_A P(A) \log_2 P(A)\ dA \qquad (3)$$

H(A) is defined as the uncertainty before B is known. H(A/B) is defined as the uncertainty after B is known.
In the equation 2, the function p(A, B) is defined as the joint probability function and P(A), P(B) as the marginal probabilities. In our proposed approach, the mutual information will be used to evaluate the discriminative ability of the bands to the classification by measuring the amount of information that each band can bring to the classification. The Bands with zero MI factor will reflects that they are independent. While the bands with high MI factor with the ground truth are considered as the most informative and discriminative one.

b) The average of interaction information controls simultaneously the redundancy and the interaction between bands in order to increase the classification. In order to control the redundancy and the information interaction simultaneously, we introduced the information gain term to our goal function (equation 4). This factor is defined by the interaction information coefficient of the ground truth, the ground truth estimated and the candidate band divided by the cardinality as below:

$$Information\ Gain\ (B) = 1/S * I(GT, GTest, B) \qquad (4)$$

Where B is the candidate band, GT is the ground truth or class label and GTest is the ground estimated that we will detail in the next part the way to calculate it.
The interaction information I(A;B;C) introduced in the information gain term has been defined by Jakulin[20] for the first time as the decrease in uncertainty caused by joining attributes A and B in a Cartesian product. The equation 5 illustrate the relation between mutual info and the interaction information measure as below:

$$I(A,B,C) = I(A;B,C) - I(A,C) - I(B,C) \qquad (5)$$

The measure value is the same, even when the class label C is swapped with one of the features A or B in Equation (5). Within the feature selection researches, filter selection using interaction information was proposed by Elmaizi [21] and Akadi [22].
In the proposed approach, the interaction information is positive when two bands together provide information about the ground truth which cannot be provided by each of them individually; it is negative when each of them can provide the same information, and zero when the two bands are independent in the context of the class label.
The ground truth estimated is generated in each step using the average of the already selected bands with the new candidate band.

The proposed approach of band selection is as follows:

1. **Initialization:**
   Initialize the selected bands subset S with the band Bi that have the largest mutual information value with the GT $Argmax\ (I(B_i, GT))$.
   This band Bi is used to create the ground truth estimated as a first step.

2. **Bands subset creation:**

   - Compute the mutual information between the candidate band and the GT : I (Bi, GT)    with (B$i$ ∈ B).

   - Compute the information gain using the mean of the interaction information between the candidate band (Bi), the GT and the estimated $GT_{est}$: 1/S* I ($GT_{est}$, Bi, GT)    with (B$i$ ∈B).

   - During each step select the band B that maximizes the information gain criterion.

3. **Output the subset S containing the selected bands**

## 3. Datasets, results and analysis

*3.1. Experimental Datasets*

The experiment has been carried out on two publicly available and widely used hyperspectral images: the Indian Pines AVIRIS and the Pavia University.

The Indiana pines dataset shown in Fig. 1, was originally acquired in June 1992 over an agricultural site composed of agricultural fields with regular geometry and with a variety of crops. It contains 220 images taken on the region "Indiana Pine" at "north-western Indiana", USA. The 220 called bands are taken between 0.4m and 2.5m. Each band has 145 lines and 145 columns. The ground truth map is also provided, but only 10366 pixels are labeled from 1 to 16 presented in figure 1. Each label indicates one from 16 classes. Each pixel of the ground truth map has a set of 220 numbers (measures) along the hyperspectral image. This dataset represents a very challenging land-cover classification problem dominated by similar spectral classes and mixed pixels.

The Pavia University image was gathered by the Reflective Optics System Imaging Spectrometer (ROSIS) optical sensor over the Pavia University, Italy. This image is 610 × 340 pixels, with a spatial resolution of 1.3 m/pixel. The ROSIS sensor generates 115 bands in the range of 0.43−0.86 μm, where 12 bands are identified as noisy bands. Only the data without noisy bands can be obtained. There are nine categories in the Pavia University image, as shown in figure 2. A three-band false color image and its ground truth map are illustrated in figure 2 (a) and (b), respectively. It covers an urban environment, with various solid structures, natural objects.

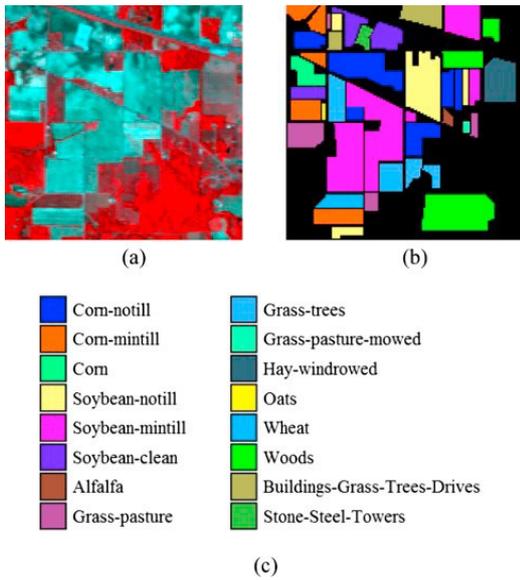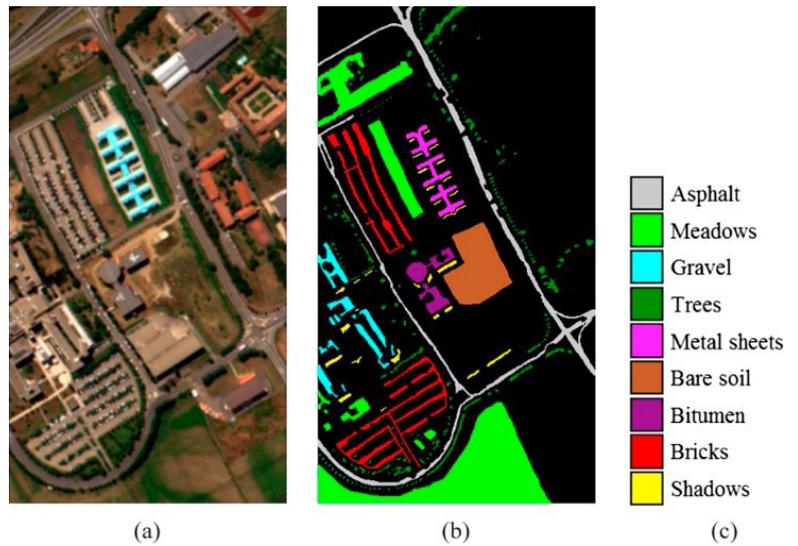

Figure 1.(a) Three-band color composite of the Indian Pines image (bands 43, 30, and 21); (b) and (c) ground truth data of the Indian Pines image.

Figure 2.(a) Three-band color composite of the University of Pavia image (bands 56, 33, and 13); (b) and (c) ground truth data of the University of Pavia image.

*3.2. Parameter setting*

In this study, the classification accuracy rate was calculated using the Support Vector Machine [17]. It is a supervised classification method based on structural risk minimization. The key idea of this technique is to estimate a separator boundary or surface between the spectral classes. This surface [18], which maximizes the margin between classes, uses limited numbers of boundary pixels (support vectors) to create the decision surface. The SVM has been compared to other classification methods for remote sensing imagery, such as the neural networks, nearest neighbor, maximum likelihood, and decision tree classifiers, and have surpassed them all in robustness and accuracy for remote sensing data. 50% labelled pixels are randomly selected to be used in training, and the other 50% will be used for the classification testing.

To evaluate the performance of the proposed approached, the overall accuracy and kappa metrics will be used to measure the classification of the proposed approach versus the state of art methods.

- The overall accuracy "OA" that reflect the percentage of the correctly classified pixels.

- The kappa coefficient is the percentage of correctly classified pixels that are expected purely by chance.

*3.3. Results and analysis*

The following table I presents the results obtained using the proposed method based on the information gain. The overall classification accuracy and the kappa indices are calculated for 80 selected band.

The classification accuracy for each one of the 16 class is presented in table 1.

The proposed approach is compared with the MRMR, MIFS and MIFB algorithms presented in the related work part.

Table 1. Results of the proposed IGBS algorithm and comparison with the state-art algorithms for Indiana pines

| Class | MIFB | MIFS | MRMR | IGBS |
|---|---|---|---|---|
| Alfafa | 78.26 | 78.26 | 78.26 | **86.96** |
| Corn-no till | 78.24 | 80.47 | 81.03 | **86.33** |
| Corn-min till | 85.85 | 76.26 | 75.54 | **82.97** |
| Corn | 69.23 | 71.79 | 73.50 | **82.05** |
| Grass/pasture | 92.68 | 92.68 | 93.50 | **93.90** |
| Grass/trees | 96.37 | 97.21 | 97.21 | **98.60** |
| Grass/pasture-mowed | 76.92 | 76.92 | 69.23 | **84.62** |
| Hay-windrowed | 97.14 | 97.96 | 98.37 | **98.37** |
| Oats | 100 | 40.00 | 50.00 | **60.00** |
| Soybeans-no till | 85.74 | 79.75 | 80.99 | **89.26** |
| Soybeans-min till | 87.93 | 86.55 | 87.84 | **90.92** |
| Soybeans-clean till | 88.27 | 85.34 | 85.99 | **88.27** |
| Wheat | 98.06 | 98.06 | 98.06 | **98.06** |
| Woods | 95.21 | 96.60 | 96.45 | **97.06** |
| Buildings-grass-tree-drives | 60.84 | 71.08 | 71.08 | **75.30** |
| Stone-steel towers | 91.30 | 91.30 | 91.30 | **93.48** |
| Kappa(%) | 92.82 | 92.76 | 93.05 | **94.94** |
| OA(%) | 93.27 | 93.21 | 93.49 | **95.25** |

Table 2. Results of the proposed IGBS algorithm and comparison with the state-art algorithms for Pavia University

| Class | MIFB | MIFS | MRMR | IGBS |
|---|---|---|---|---|
| Asphalt | 90.93 | 96.02 | 95.99 | **96.08** |
| Meadows | 97.36 | 97.71 | 97.75 | **98.64** |
| Gravel | 67.78 | 80.98 | 81.17 | **85.56** |
| Trees | 94.40 | 95.85 | 95.85 | **97.43** |
| Metal sheets | 100 | 99.85 | 99.85 | **100** |
| Bare soil | 84.24 | 89.09 | 88.93 | **93.35** |
| Birumen | 74.55 | 90.00 | 90.45 | **90.15** |
| Bricks | 86.79 | 91.98 | 91.98 | **93.12** |
| Shadows | 100 | 99.79 | 99.79 | **99.79** |
| Kappa(%) | 91.68 | 95.07 | 95.10 | **96.43** |
| OA(%) | 92.60 | 95.62 | 95.64 | **96.83** |

The above table (I) and table (II) shows the strength of our proposed approach compared with mutual information selection filters (MIFS, MIFB, MRMR).
The analysis of the previous tables I and II allowed us to notice the following results:
The information gain-based selection filter (IGBS) proposed approach selects bands with high discrimination power very quickly. On the Indiana pines, IGBS achieves 95.25% classification accuracy with 80 bands, which is higher than the closest method (MRMR) by 2% and higher by more than 2.5% % from the MIFS and the MIFB. For the Pavia university dataset, the proposed approaches achieve 96.83% which is higher than the MIFB with more than 4%.
From figures 3 & 4, we detect that reducing the datasets to 80 bands are sufficient to detect all the materials in different classes from the ground truth estimated that we've generate.
These algorithms performs better than the ranking mutual information algorithm (MIM) that is only based on selecting the relevance bands based only on mutual information with the ground truth. This result explains the benefit of adding the redundancy control term to the goal function used by both algorithms and which leads to achieve a better classification using the same number of selected bands. The results also shows that MRMR perform better than the MIFS approaches because the redundancy term is divided over the cardinality of the subset in order to avoid the confusion that may happens when the subset becomes very large. The redundancy term in this case can be very large compared to the relevance term and automatically the algorithm may select irrelevant bands because they are not redundant but not because they are relevant to the ground truth, which will lead to the classification rate decreasing.

The algorithm MIFB reproduced previously is based on feature selection using MI to select the bands that are able to classify the pixels of the Ground truth. Despite its simplicity and rapidity, it represents several drawbacks such as using the threshold that may lead to lose the selection (we used TH=-0.02 to generate the results).

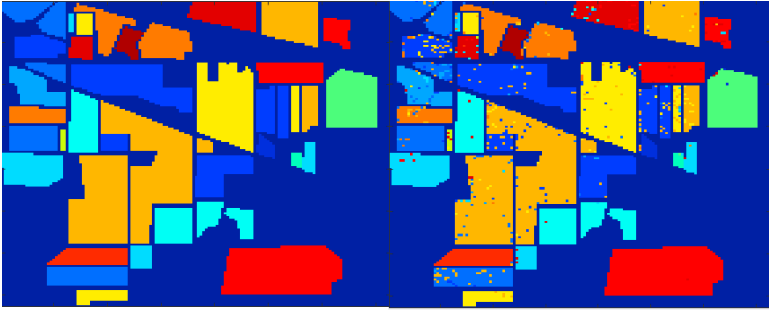

Figure 3. Indiana Ground truth (left), Indiana Ground truth estimated (right) using the proposed algorithm for 80 bands.

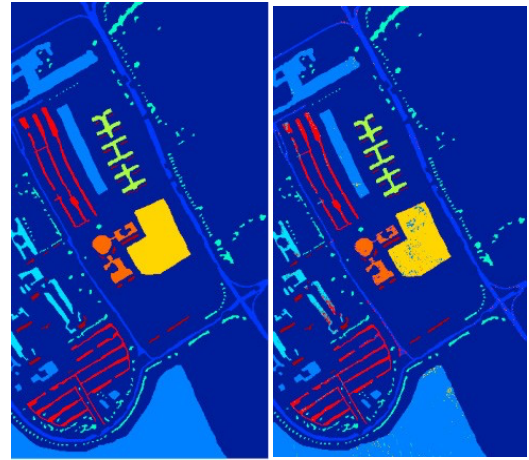

Figure 4. Pavia Ground truth (left), Pavia Ground truth estimated (right) using the proposed approach for 80 bands.

The algorithms based on mutual information eliminates irrelevant bands based on their correlation with the GT, without thinking that, they can be very important for discriminating the ground truth when combined with other bands. These approaches neglected the information gain between the selected bands. It is also worth noting that IGBS performs better than the methods which are not based on interaction information. Which explains the importance of the interaction information as a measure for excluding redundant, irrelevant bands from the subset and at the same time guarantee the complementarity between the selected band subsets.

## 4. Conclusion and perspectives

This paper presents a new filter selection approach based on information gain in order to deal with the dimensionality reduction and classification of hyperspectral images.
The proposed method selects relevant, non-redundant and discriminative bands by testing the interaction and the complementarity between the candidate bands and the bands already selected within the subset. The proposed approach is compared with three other selection algorithms using the SVM classifier and conducting the experiment on the AVIRIS and PAVIA image datasets provided by the NASA.
The results shows that the proposed method based on the information gain criterion outperforms the reproduced algorithms. Its major advantage is the redundancy, relevance and complementarity simultaneously. We aim to evaluate the performance of our approach using other image datasets and classifier as well as studying the time complexity that will be a major point to develop for our future works and study.